\pdfoutput=1

\documentclass[conference]{IEEEtran}
\ifCLASSINFOpdf
\else
\fi
\usepackage[utf8]{inputenc} 
\usepackage[T1]{fontenc}    
\usepackage[colorlinks,linkcolor=red,anchorcolor=blue,citecolor=green]{hyperref}       
\usepackage{url}            
\usepackage{booktabs}       
\usepackage{amsfonts}       
\usepackage{nicefrac}       
\usepackage{microtype}      
\usepackage{amsmath,amssymb}
\usepackage{graphicx}
\usepackage{pifont}
\usepackage{enumitem}
\usepackage{graphicx}
\usepackage{booktabs}
\usepackage{multirow}
\usepackage[table]{xcolor}
\usepackage{siunitx}
\usepackage{pifont}
\usepackage{newunicodechar}
\usepackage{algpseudocode}
\usepackage{subcaption} 
\usepackage[a4paper,margin=1in]{geometry}
\usepackage{caption}       
\usepackage{makecell}

\usepackage{cite}
\usepackage{algorithm}

\usepackage{etoolbox}

\begin{document}

\title{SMA: Who Said That? Auditing Membership Leakage in Semi-Black-box RAG Controlling}

\author{\IEEEauthorblockN{Anonymous Author(s)}
	\IEEEauthorblockA{Affiliation\\
		email}
	}

\author{\IEEEauthorblockN{Shixuan Sun}
	\IEEEauthorblockA{Sun Yat-Sen University\\
		ssun0526@uni.sydney.edu.au}
	\and
	\IEEEauthorblockN{Siyuan Liang}
	\IEEEauthorblockA{Nanyang Technological University\\
		siyuan.liang@ntu.edu.sg}
    \and
    \IEEEauthorblockN{Jianjie Huang}
	\IEEEauthorblockA{Zhongguancun Academy\\
		huangjj67@mail2.sysu.edu.cn}
    \and
    \IEEEauthorblockN{Jingzhi Li}
	\IEEEauthorblockA{Institute of Information Engineering, \\ Chinese Academy of Sciences\\
		lijingzhi@iie.ac.cn}
    \and
    \IEEEauthorblockN{Xiaochun Cao\IEEEauthorrefmark{1}}
    \IEEEauthorblockA{Sun Yat-Sen University\\
    caoxiaochun@mail.sysu.edu.cn}
    \thanks{\IEEEauthorrefmark{1}Corresponding author.}}

\maketitle

\setlength{\parskip}{0pt}
\setlength{\parindent}{2em}

\begin{abstract}

Retrieval-Augmented Generation (RAG) and its Multimodal Retrieval-Augmented Generation (MRAG) significantly improve the knowledge coverage and contextual understanding of Large Language Models (LLMs) by introducing external knowledge sources. However, retrieval and multimodal fusion obscure content provenance, rendering existing membership inference methods unable to reliably attribute generated outputs to pre-training, external retrieval, or user input, thus undermining privacy leakage accountability

To address these challenges, we propose the first \underline{S}ource-aware \underline{M}embership \underline{A}udit (SMA) that enables fine-grained source attribution of generated content in a semi-black-box setting with retrieval control capabilities.
To address the environmental constraints of semi-black-box auditing, we further design an attribution estimation mechanism based on zero-order optimization, which robustly approximates the true influence of input tokens on the output through large-scale perturbation sampling and ridge regression modeling.
In addition, SMA introduces a cross-modal attribution technique that projects image inputs into textual descriptions via MLLMs, enabling token-level attribution in the text modality, which for the first time facilitates membership inference on image retrieval traces in MRAG systems.
Experiments on multiple textual and multimodal RAG benchmarks show that SMA outperforms state-of-the-art black-box MIA baselines in detecting source-specific membership leakage, with notable improvements in accuracy (+15.74\%) and coverage metrics (+10.01\%) under noise and zero-gradient conditions, demonstrating the effectiveness of attribution-based strategies in differentiating content origins.
This work shifts the focus of membership inference from ``whether the data has been memorized'' to ``where the content is sourced from'', offering a novel perspective for auditing data provenance in complex generative systems.

\end{abstract}

\IEEEpeerreviewmaketitle

\section{Introduction}
Large Language Models (LLMs) and Multimodal Large Language Models (MLLMs) have made significant progress in natural language understanding and generation tasks. With the rise of Retrieval-Augmented Generation (RAG) and Multi-model Retrieval-Augmented Generation (MRAG), models can dynamically access external knowledge bases during inference, enabling them to supplement responses with up-to-date or domain-specific information. While this improves the relevance and accuracy of responses, it also raises privacy and safety risks~\cite{zhou2025semidefiniteprogrammingrelaxationsdebiasing, luo2024reactfaceonlinemultipleappropriate，liang2023badclip,liang2025revisiting,lu2025adversarial,liang2025t2vshield,ying2024jailbreak,ying2025pushing,liang2022imitated,li2023privacy,xiao2024genderbias,xiao2025fairness}. For example, external sources may contain sensitive or proprietary data that can be exported by the model, such as corporate documents, support tickets, or patient histories\cite{10.1145/3637528.3671470}, leading to accidental content disclosure.

To cope with the privacy leakage problem in generative models, Membership Inference Attacks (MIA)~\cite{wu2025membershipinferenceattackslargescale}~\cite{hu2022membershipinferenceattacksmachine}~\cite{bai2024membershipinferenceattacksdefenses} are widely used to identify whether a particular data instance has been used during model training.
However, in RAG systems, existing techniques struggle to accurately infer the origin of generated content and face the following two main challenges: (1) Destabilized input-output associations. RAG/MRAG systems dynamically incorporate externally retrieved content at inference time, which is concatenated with the original query and passed into the model to produce responses. This type of input augmentation and blending disrupts the stable input-output correspondence that traditional MIA methods rely on, making it difficult for the attacker to pinpoint which parts of the output stem from the original query. (2) Unobservable influence paths due to multimodal fusion. In MRAG, images are encoded into latent feature representations before being consumed by the model, and these are jointly processed with text to guide generation. Since the attacker cannot directly access or inspect the retrieved image content, and the model output integrates signals from multiple modalities, the contribution path of potential leakage becomes opaque.

\begin{figure*}
    \centering
    \includegraphics[width=1\linewidth]{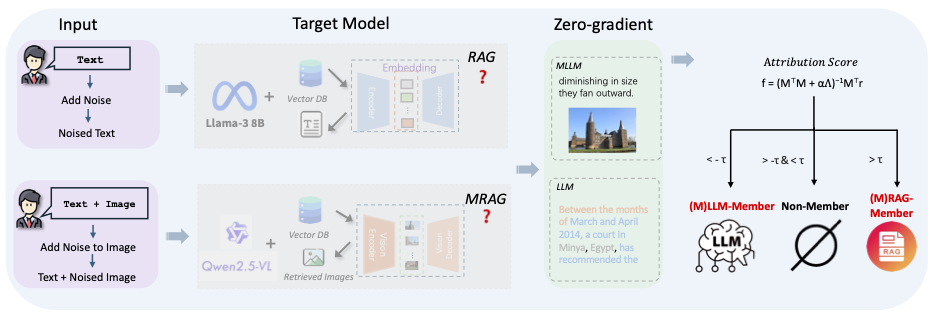}
    \caption{An example of LLM/RAG MIA}
    \label{fig:introl}
\end{figure*}

To address the difficulty of tracing the origin of generated content, this paper proposes SMA, the first source-aware membership inference framework for RAG and MRAG systems, which, unlike the traditional MIA approach that only determines ``whether or not it is memorized,'' can further identify whether the leaked content originated from the model's pre-training corpus or from external retrieval results. This capability is important for enabling content traceability and privacy risk assessment.

SMA operates in a semi-black-box environment, it does not have access to the internal structure of the model or the gradient information, but it has control over the switching of the retrieval module. In realistic deployment scenarios, enabling or disabling the RAG component often hinges on the application's requirements: when response timeliness is crucial, the retrieval mechanism might be turned off to ensure faster inference; conversely, when high knowledge accuracy and confidence in responses are prioritized, the RAG mode is typically activated to incorporate external, up-to-date information. However, it is essential to acknowledge that not all commercial platforms explicitly expose this toggle functionality—for instance, Grok currently lacks this direct user-controllable option.

Figure~\ref{fig:introl} illustrates the overall flow of SMA, which is designed to capture source bias in the output behavior by constructing lightweight input perturbations and toggling the retrieval module on/off, using an inference mechanism based on attributional discrepancies. Meanwhile, we introduce a zero-gradient scoring strategy to estimate the influence of input tokens on outputs using large-scale perturbation sampling with ridge regression modeling without gradient access. For compatibility with multimodal inputs, SMA also builds a unified cross-modal attribution mechanism to transform images into textual descriptions via MLLM outputs with token-level attribution in the textual domain, thus supporting for the first time membership inference of image retrieval paths in MRAG.

Experiments on multiple textual and multi-model RAG systems demonstrate that SMA significantly outperforms state-of-the-art membership inference approaches in source-specific leakage auditing. SMA achieves 15.74\% improvement in metric ACC and 10.01\% in metric Coverage under $\text{top}_{\text{k}}$ retrieval settings, confirming its effectiveness in fine-grained provenance analysis. Our contributions are summarized as follows: 
(1) We introduce the source-aware membership auditing problem, which extends traditional membership inference to differentiate between internal and external content sources in RAG and MRAG systems.
(2) We propose SMA, a semi-black-box auditing framework that performs fine-grained source attribution via input perturbation and a novel zero-gradient scoring mechanism.
(3) Experimental results on multiple RAG systems show that SMA achieves superior performance in source leakage auditing, improving ACC by 33.94\% and Coverage by 25.32\% over the existing baseline.

\section{Related Work}
\label{sec: Related Work}

\subsection{Membership Inference Attacks}
MIAs aim to determine whether a given data point was included in the training set of a target model by observing its behavior on that point. Early work by Shokri \emph{et al.}~\cite{7958568} demonstrated practical MIAs on classification models via the shadow models and posterior probabilities \cite{pmlr-v139-choquette-choo21a}.  S. Yeom \emph{et al.}~\cite{8429311} further showed that only label access can suffice when combined with loss thresholds and sensitivity measures.

More recently, there has been more attention on RAG~\cite{Anderson_2025}~\cite{liu2025maskbasedmembershipinferenceattacks}, Huang et al.\cite{huang2023privacy} demonstrated that RAG systems are more likely to output verbatim sensitive strings from their retrieval databases, especially when attackers craft adversarial prompts. Zeng et al.\cite{zeng-etal-2024-good} further expanded on this by demonstrating privacy leakage from real-world datasets such as Enron emails and medical records, and even proposed automated prompt injection attacks that can extract raw sensitive content with high precision. Compounding this issue, optimization-based attacks such as DEAL\cite{zhang2025deal} have shown that attackers can automatically generate prompts which can maximize privacy leakage from RAG systems. Naseh \emph{et al.}~\cite{nasehRiddleMeThis2025} proposed the \emph{Interrogation Attack}, crafting natural‐language queries that are answerable only if a target document resides in the RAG vector database, achieving high inference accuracy and remaining stealthy against prompt‐injection detectors.  Similarly, the RAG-Thief~\cite{jiangRAGThiefScalableExtraction2024} generates probing prompts to elicit context membership without triggered safe guarding. 

Despite these advances, important gaps remain.  First, no existing work considers MIA against multi-modal RAG (MRAG) systems combined with MLLMs. Second, current RAG MIAs cannot accurately attribute retrieved information to either the external database or the LLM’s internal knowledge, leading to decreased accuracy when RAG is enabled on LLMs. Our work fills these gaps by proposing the first source-aware MIA for MRAG\&MLLM semi-black-boxed system that both detects image membership in the retrieval set and discriminates between Retrieved Member and Pretrained Member.

\subsection{Retrieval-Augmented Generation}

RAG enhances a base language model by consulting an external text datastore~\cite{khan2024developingretrievalaugmentedgeneration} at inference time. Given a text query, a dense or sparse retriever (\textit{e.g., DPR~\cite{karpukhin-etal-2020-dense} or BM25~\cite{inproceedings}}) fetches the most relevant requirements, which are then prepended to the query and fed into a generative model. This design enables on-demand access to up-to-date or domain-specific knowledge without retraining or fine-tuning the model.

MRAG extends RAG to incorporate image data alongside text. In MRAG, two main retrieval modes coexist, the first is Text-to-Image Retrieval where a text query is projected into a joint embedding space (\textit{e.g., via CLIP~\cite{radford2021learningtransferablevisualmodels}}) to retrieve semantically matching images. The second is Image-to-Image Retrieval~\cite{song2025comprehensivesurveycomposedimage} where a visual query is used to fetch related images that provide additional context for generation.
The selected images are encoded by a vision encoder and fused with text embeddings to prompt a multi-model generative model.

Considering RAG and MRAG rely on external datastores that can be updated or pruned independently of the model, they are both attractive targets for leaking or poisoning sensitive data (\textit{e.g.\ private medical records, proprietary corporate documents}).  Unlike monolithic LLMs, where forgetting requires expensive fine-tuning~\cite{zhang2024instructiontuninglargelanguage}, removing or auditing entries in a vector database is relatively easy—but often overlooked. Moreover, an attacker could insert malicious or private content into the datastore, enabling stealthy membership inference or data exfiltration attacks~\cite{Chen_2020}~\cite{liu2019performingcomembershipattacksdeep}. Thus, we argue that the security of RAG/MRAG knowledge bases—encompassing fine-grained access control, verifiable deletion logs, and adversarial filtering—is paramount for safe deployment in sensitive domains.

\subsection{Black-Box Attribution Methods}

The attribution methods for generative models~\cite{cai2024gradientlikeexplanationblackboxsetting}~\cite{zhao2024reagentmodelagnosticfeatureattribution}~\cite{Id__2023}~\cite{zaher2024manifoldintegratedgradientsriemannian}~\cite{simpson2025tangentiallyalignedintegratedgradients} seek to explain which parts of the input are most responsible for certain outputs, offering valuable insights into model behavior without requiring full access to internal parameters. Early model-agnostic approaches such as LIME~\cite{ribeiro2016whyitrustyou} fit simple surrogate models around individual predictions by perturbing inputs and observing changes in outputs, thereby estimating feature importance through local linear approximations. SHAP (Shapley additive explanations)~\cite{NIPS2017_8a20a862} further casts this perturbation-based reasoning into a principled game-theoretic framework, attributing output differences to input features via Shapley values.

In our work, we leverage both black-box attribution techniques to distinguish whether generated content from an external RAG database or from the model’s pre-trained parameters. By comparing the attribution signatures under these two settings, we are able to infer the token-level source, achieving a membership inference in both black-box RAG\&LLM and MRAG\&MLLM systems.  
\section{Background} 

\label{sec:threadmodel}
\subsection{Preliminary}

\textbf{Inputs and model definition.} To establish the modeling basis for subsequent attribution mechanisms, we first define the input format of the respective LLMs~\cite{10.1145/3639372} and MLLMs~\cite{Yin_2024}~\cite{zhangMMLLMsRecentAdvances2024}~\cite{mei2025surveymultimodalretrievalaugmentedgeneration}.

In a pure text scenario, an LLM receives an input token sequence \(\mathbf{x}^{\text{text}} = (w_1, w_2, \dots, w_k)\), where each token \(w_i \in \mathcal{V}\), and \(\mathcal{V}\) denotes the model's vocabulary. The model generates an output text \(\mathbf{y}\) from this input, denoted as:
\begin{equation}
    \mathbf{y} = f(\mathbf{x}^{\text{text}})
\end{equation}

In a multimodal scenario, an MLLM not only receives text but also visual inputs such as images. The image \(\mathbf{I}\) is converted into a visual token sequence via a visual encoder (\textit{e.g., ViT}):
\begin{equation}
    \mathbf{v}^{\text{vis}} = \mathrm{VisionEnc}(\mathbf{I}).
\end{equation}

These visual tokens are then concatenated with the text tokens \(\mathbf{x}^{\text{text}}\) to form the input sequence:
\begin{equation}
  \mathbf{x}^{\text{multi}} = \texttt{concat}(\mathbf{v}^{\text{vis}},\ \mathbf{x}^{\text{text}})  
\end{equation}
The corresponding generated output denotes \(\mathbf{y} = f(\mathbf{x}^{\text{multi}})\).

\textbf{Retrieval-augmented generation}. 
To enhance the knowledge and contextual coverage of generated content, RAG and its MRAG introduce external retrieval modules that enable models to dynamically incorporate additional evidence from textual or visual sources. We introduce the input modeling of text-based RAG and image-based MRAG separately, as their retrieval modalities and token representations differ.

\emph{Text-to-Text RAG}. 
Given a sequence of text tokens \(\mathbf{x}^{\text{text}}\) as user input, a retriever queries an external document corpus \(\mathcal{D}^{\text{text}}\) to obtain the top-\(k\) most relevant passages:
\begin{equation}
\mathbf{x}^{\text{retr}} = \{d_1, \dots, d_k\}.
\end{equation}

These retrieved passages are appended to the original input, forming the final input sequence:
\begin{equation}
\mathbf{x}^{\text{RAG}} = \texttt{concat}(\mathbf{x}^{\text{text}},\ \mathbf{x}^{\text{retr}}).
\end{equation}

The model then generates the output based on this combined context:
\begin{equation}
\mathbf{y} = f(\mathbf{x}^{\text{RAG}}).
\end{equation}

Formally, the model~\cite{kaplan2020scalinglawsneurallanguage} computes the conditional distribution over the next token \(y_t\) autoregressively, conditioned on the previous tokens and the full context:
\begin{equation}
 P(\mathbf{y} \mid \mathbf{x}^{\text{text}},\ \mathbf{x}^{\text{retr}}) =\prod_{t=1}^T P(y_t \mid y_{1:t-1},\ \mathbf{x}^{\text{text}},\ \mathbf{x}^{\text{retr}}).
\end{equation}

\emph{Image-to-Image MRAG}. 
In MRAG settings, the model receives both an original image \(\mathbf{I}_{\mathrm{orig}}\) and an optional text prompt \(\mathbf{x}^{\text{text}}\). We use a CLIP-based image encoder to embed the input image and perform nearest-neighbor retrieval from an external image corpus, returning the $\text{top}_\text{k}$ most similar images \(\{\mathbf{I}'_1,\dots,\mathbf{I}'_k\}\). Each image is transformed into a sequence of visual tokens:

\begin{equation}
\begin{split}
\mathbf{v}^{\text{retr\_vis}} &= \{\mathrm{VisionEnc}(\mathbf{I}'_i)\}_{i=1}^k.
\end{split}
\end{equation}

The final input to the MLLM is constructed by concatenating the retrieved visual tokens, the original image tokens, and the textual prompt:
\begin{equation}
\mathbf{x}^{\text{MRAG}} = \texttt{concat}(\mathbf{v}^{\text{retr\_vis}},\ \mathbf{v}^{\text{vis}},\ \mathbf{x}^{\text{text}}).
\end{equation}

Based on this multimodal context, the model produces the output:
\begin{equation}
\mathbf{y} = f(\mathbf{x}^{\text{MRAG}}).
\end{equation}

\subsection{Threat Model}
\textbf{Audit target systems}.
We focus on two representative deployment scenarios as audit targets: RAG\&LLM and MRAG\&MLLM systems.

In RAG\&LLM systems, the model receives a text query and augments it with ${top_k}$ retrieved text passages from an external corpus. These passages are concatenated with the original input and fed into the language model for generation.

In MRAG\&MLLM systems, the input includes both an image and optional text. The system retrieves related images from an external image corpus based on visual similarity, and encodes both the original and retrieved images through a modality interface before passing them into a multimodal language model.

These retrieval-augmented designs enhance response quality but also blur the boundary between model-internal knowledge and external sources, complicating attribution and raising new challenges for membership auditing.

\textbf{Auditing goal}.
Traditional MIA aim to determine whether a given input sample was included in a model’s training corpus. However, in RAG and MRAG systems, generated content may originate from either the model’s pretraining data or external retrieved content. Therefore, we extend the auditing objective to a three-way classification task: for any input \(\mathbf{x}\), determine whether it is: (1) \emph{Pretrained Member}: The sample was encountered by the model during pretraining; (2) \emph{Retrieved Member}: The sample exists in the external retrieval database and is leaked via RAG/MRAG; (3) \emph{Non-Member}: The sample does not appear in either the pretraining corpus or the external retrieval database.
This auditing task not only determines whether the generated content constitutes a data leakage event, but also identifies the leakage pathway—\textit{i.e}., whether the source is the model itself or an external resource.

\textbf{Audit capability settings}.
We assume the auditor possesses the following capabilities: (1) \emph{Black-box access} to the target system, including the ability to provide text and image inputs and observe corresponding outputs; (2) \emph{Retrieval control}, \textit{i.e}., the ability to toggle RAG or MRAG modules on or off (\textit{e.g., to control whether retrieval is performed}); (3) \emph{Input perturbation}, \textit{i.e}., the ability to apply fine-grained modifications to inputs in order to construct counterfactual examples that elicit divergent outputs.
Without access to model weights, gradients, or attention mechanisms, we design an attribution mechanism to determine the provenance path of the generated content, thereby enabling cross-modal, multi-source membership auditing.

\section{Method} 
\label{sec: Preliminary}

\subsection{Problem Definition}

In the context of the aforementioned generation system, the audit issue we are concerned with is not the traditional question of ``whether the output contains a certain piece of data,'' but rather a more fine-grained determination: given an input query (or its internal key substrings), does the information originate from the models' pretraining corpus, external retrieval repository, or novel content directly provided by the users' input?

Specifically, the audit task can be formalized as follows: Given an input \(\mathbf{x}\) (comprising text or images), and its corresponding black-box model $f_\text{black}$ (the LLM/MLLM with or without RAG systems), determine whether the input falls into one of three categories. This distinction not only determines ``whether leakage has occurred,'' but also further identifies the ``leakage pathway.'' We denote this three-way classification task as:

\begin{equation} \textit{}
 \mathcal{A}(f_\text{black}, \mathbf{x}) \rightarrow 
\left\{
\begin{array}{ll}
\texttt{Pretrained Member},\\
\texttt{Retrieved Member}, \\
\texttt{Non-Member}.
\end{array}
\right.
\end{equation}

Since modern systems are often deployed in RAG or MRAG configurations with controllable switches, we assume that auditors possess the following minimum capabilities: (1) the ability to provide inputs to the system and observe corresponding outputs (black-box access); (2) the ability to toggle the retrieval module on or off; (3) the ability to apply controlled perturbations to the input to construct counterfactuals (\textit{e.g., key substring substitution, Gaussian noise perturbation on images, etc.}). The objective is to determine, under this setting, whether the model’s response is based on internally acquired knowledge or externally retrieved content solely through input-output interaction, thereby enabling attribution analysis of the membership relationship between the model’s pretraining corpus and retrieval sources.

\subsection{Input Perturbation Design}
To induce observable changes in the model's generative behavior, we first apply controlled perturbations to the original input. These perturbations specifically target key semantic components within the input, aiming for ``light perturbations with high responsiveness,'' \textit{i.e}., inducing significant differences in whether the model relies on external retrieval content (RAG) or its own pre-trained knowledge (LLM) without compromising semantic integrity.

In the text modality, we employ keyword-level semantic perturbations. Defined the i-th word ${w_i}$, let the original text input be $\mathbf{x}^{\text{text}} = \{w_1, w_2, \dots, w_L\}$, where ${L}$ denotes the number of words.  We only perform perturbation operations on tokens in $\mathcal{P}_{\text{text}}(\cdot)$, including synonym substitution, random masking, and word-level Unicode alteration. The perturbed text ${\mathbf{y}}^{\text{text}}$is represented as:
\begin{equation}
\tilde{\mathbf{x}}^{\text{text}} = \mathcal{P}_{\text{text}}(\mathbf{x}^{\text{text}}),
\end{equation}
where $\mathcal{P}_{\text{text}}(\cdot)$ denotes the perturbation operator applied to the words.

In the image modality, we inject pixel-wise Gaussian noise with the perturbed function $\mathcal{P}_{\text{vis}}(\cdot)$ into the original query image to construct perturbation variants of varying magnitudes. Given the original image $\mathbf{I}_{\text{orig}}$, its perturbed version is defined as:
\begin{equation}
\tilde{\mathbf{I}} = \mathcal{P}_{\text{vis}} (\mathbf{I}) = \mathbf{I}_{\text{orig}} + \epsilon,\quad \epsilon \sim \mathcal{N}(0, \sigma^2),
\end{equation}
where $\sigma$ controls the perturbation strength. We select multiple $\sigma$ values (\textit{e.g., $\{10, 20, 30, 40\}$}) to observe whether the model's generation exhibits systematic shifts across different perturbation levels.

The core motivation behind this cross-modal perturbation design is that the autoregressive mechanism of LLMs is sensitive to token-level perturbations, while the embedding-based retrieval mechanism of RAG exhibits a certain degree of input robustness. Therefore, under the condition that key tokens are perturbed, if the generated results remain consistent, it is more likely to be RAG-driven; otherwise, it is more likely to rely on the LLM itself.

To accommodate the multimodal setting, we further transform the perturbed visual input  with a specific prompt into a textual representation using the model’s own image encoder and captioning head, denoted as $f_{\text{black}}(\cdot)$. Specifically, the noisy image $\tilde{\mathbf{I}}$ is processed by the model to yield a corresponding textual embedding or caption, denoted as $\tilde{\mathbf{y}}^{\text{vis}} = f_{\text{black}}(prompt,  \tilde{\mathbf{I}})$. This allows us to directly attribute the influence of visual perturbations within the same textual autoregressive decoding space. By analyzing the generation conditioned on $\tilde{\mathbf{y}}^{\text{vis}}$, we can assess whether the perturbations in the modality encoder propagate meaningfully into the generation process, and how they are weighted relative to retrieved contexts or prior tokens.

\subsection{Zero-Gradient Auditing Mechanism}
LLMs and MLLMs in practical applications are often closed-source APIs without access to model parameters or gradients, we propose a \emph{Zero-Gradient Attribution Mechanism} to estimate the contribution of input tokens to the output under semi-black-box conditions. The method does not rely on the internal architecture or backpropagation, but instead models attribution by analyzing the relationship between input perturbations and output variations.

We construct $N$ randomly perturbed variants $\tilde{\mathbf{x}}^{text(i)}$ and feed them into the black-box model $f_{\text{black}}(\cdot)$ to obtain the corresponding outputs $\tilde{\mathbf{y}}^{(i)} = f_{\text{black}}(\tilde{\mathbf{x}}^{text(i)})$. Each perturbed sample $\tilde{\mathbf{x}}^{text(i)}$ is associated with a binary mask vector $\mathbf{m}^{(i)} \in \{0,1\}^L$, where each element indicates whether the corresponding word is retained (1) or perturbed (0).

To quantify the impact of each perturbation on the model's output, we define an attributed response score as:
\begin{equation}
\mathbf{r}^{(i)} = \gamma_{1}\,
\frac{\operatorname{Len}\!\bigl(\tilde{\mathbf{y}}^{(i)}\bigr)}
     {\operatorname{Len}(\tilde{\mathbf{x}}^{text(i)})}
\;+\;
\gamma_{2}\,
\operatorname{Sim}\!\bigl(\tilde{\mathbf{y}}^{(i)},\,\tilde{\mathbf{x}}^{text(i)}\bigr),
\end{equation}
where $\mathrm{Len}(\cdot)$ denotes the output length, and $\mathrm{Sim}(\cdot, \cdot)$ represents a semantic similarity metric (\textit{e.g., BLEU, cosine similarity, BERTScore}). The weighting factors $\gamma_1$ and $\gamma_2$ control the contribution of each term. A lower $\mathbf{r}^{(i)}$ indicates a greater perturbation impact on the output.

We then stack all mask vectors into a perturbation matrix $M \in {0,1}^{N \times L}$ and assemble all response scores into a vector $\mathbf{r} \in \mathbb{R}^N$. To estimate token attribution, we fit a linear regression model:
\begin{equation}
\mathbf{r} = \mathbf{M} \boldsymbol{\beta},
\end{equation}
where $\boldsymbol{\beta} \in \mathbb{R}^L$ denotes the attribution scores to be estimated. We adopt ridge regression with regularization to obtain a closed-form solution:
\begin{equation}
\boldsymbol{\beta} = (\mathbf{M}^\top \mathbf{M} + \alpha\boldsymbol{\Lambda}\bigr)^{-1} \mathbf{M}^\top \mathbf{r},
\end{equation}
where $\alpha > 0$ is a ridge coefficient used to reduce multicollinearity and improve robustness, and $\boldsymbol{\Lambda}$ denotes the $L \times L$ identity matrix. Each component $\boldsymbol{\beta}_j$ thus represents the estimated contribution of the $j$-th token $w_j$ to the model's output, serving as its attribution strength. This approach performs zero-order sensitivity analysis purely based on input-output behaviors, without relying on gradient information.

In the multimodal setting, if the input includes an image $\mathbf{x}^{\text{vis}}$, we apply small Gaussian noise perturbations $\tilde{\mathbf{x}}^{\text{vis}} = \mathcal{P}_{\text{vis}}(\mathbf{x}^{\text{vis}})$ and process it via the model's visual encoder to generate a textual representation $\tilde{\mathbf{y}}^{\text{vis}} = f_{\text{black}}(\tilde{\mathbf{x}}^{\text{vis}})$, which is then integrated into the text generation pipeline. This allows image perturbations to be translated into measurable changes in the token-level textual output.

The proposed mechanism is applicable to text-only, multimodal, and retrieval-augmented generation (RAG) scenarios, and enables unified attribution analysis under semi-black-box constraints.

\subsection{Attribution Scoring under RAG Switch}
To quantitatively characterize the difference in contribution between RAG and the LLM in generating outputs, we introduce the \textit{Attribution Difference Score} (ADS), which measures the change in the impact of key words on the generated results by toggling the retrieval component. 

We generate outputs for the same perturbed input with the retrieval module enabled and disabled, respectively, and denote the model outputs as:
\begin{equation}
   \boldsymbol{\beta}_{\text{RAG}} = Ridge({\mathbf{x}^{text},\hat{y}};\ \hat{\mathbf{y}}_\text{(M)RAG})
\end{equation}
\begin{equation}
\boldsymbol{\beta}_{\text{w/o RAG}} = Ridge(\mathbf{x}^{text},\hat{y};\ \hat{\mathbf{y}}_\text{w/o (M)RAG}) 
\end{equation}

Where $\hat{\mathbf{y}}_\text{(M)RAG}$ and $\hat{\mathbf{y}}_\text{w/o (M)RAG}$ denote the outputs of $\mathbf{x}^\text{text}$ in the black-box system with (M)RAG enabled and disabled, respectively. For each word, we compute the corresponding ridge regression coefficient. For each word, the ridge regression coefficient \( \boldsymbol{w}_{j} \) represents the gradient attribution score of the \( j \)-th word \( w_{j} \), which directly indicates the sensitivity of the model's output to this word. To categorize the source of each word, we define thresholds according to the following criteria:

\begin{equation}
\text{label}(w_j) = \begin{cases}
\text{\texttt{Pretrained Member}}, & \boldsymbol{\beta}_{j} \geq \tau, \\
\text{\texttt{Retrieved/Non Member}}, & \boldsymbol{\beta}_{j} < \tau,
\end{cases}
\end{equation}

Then, we compute the modified Attribution Difference Score (ADS) for each critical word  using:
\begin{equation} \label{eq:whitediff}
Diff(\boldsymbol{\beta}_j) = {\boldsymbol{\beta}_\text{(M)RAG}}_\text{(j)} - {\boldsymbol{\beta}_\text{w/o (M)RAG}}_\text{(j)}
\end{equation}

Here,  $\text{w/}$ and $\text{w/o}$ represent outputs with and without RAG, respectively. The ADS directly measures how significantly toggling RAG affects the generated outputs, reflecting the token's dependence on retrieval content.   

Here, \( \tau \) represents the threshold distinguishing words originating from the model's pretrained knowledge from those retrieved externally. Subsequently, we further dissect the words labeled as \emph{Retrieved/Non Member} by setting additional empirical thresholds \(\tau_1 = 0.1\) and \(\tau_2 = -0.1\), applying the attribution difference score \(\text{Diff}(\boldsymbol{\beta}_{j})\) as follows:

\begin{equation}
\text{label}(w_j) = \begin{cases}
\text{\texttt{Non-Member}}, & \hspace{-3em} \tau_1 \leq \text{Diff}(\boldsymbol{\beta}_{j}) < \tau_2, \\
\text{\texttt{Retrieved Member}}, & \text{otherwise}.
\end{cases}
\end{equation}

Typically originating from user input prompts, from words retrieved via external knowledge bases. It is crucial to emphasize that the selected words, classified through this multi-tiered thresholding approach, constitute meaningful keywords, deliberately excluding trivial or semantically negligible words such as prepositions and conjunctions (e.g., \emph{and}, \emph{is}, \emph{or}). This targeted keyword approach ensures the keyword will be detected in the output of LLMs.

The core idea of this \textit{bias-based attribution} strategy is as follows: under the condition that a key word is perturbed, if disabling RAG causes a significant output shift while enabling RAG mitigates this deviation, it implies that the model’s generation depends on externally retrieved content.

\subsection{SMA Framework}
The SMA framework is designed to address the issue of knowledge source attribution in black-box settings, determining whether a particular piece of information in the model's output originates from the pre-training corpus of a LLM or from the retrieved content of an external retrieval module (RAG/MRAG). This method is based on the concept of zeroth-order attribution, which infers the importance of words or regions solely through changes in input-output pairs, even when gradients, attention weights, or internal representations are inaccessible. Specifically, SMA applies structured perturbations to the input content (text or image): in text, it randomly samples and retains a subset while injecting character-level Unicode variants; in images, it adds Gaussian noise that preserves high-level semantics. These perturbations do not affect the retrieval results but distort the model's internal encoding, enabling the inference of the input components primarily responsible for the generated output.

After collecting outputs from multiple perturbed inputs, SMA constructs response scores based on the output length and its semantic similarity to the original output. The perturbation masks and response scores are then fed into a ridge regression model to obtain token-level attribution scores. Higher scores indicate that the corresponding word has a stronger influence on the model's output. Furthermore, in multimodal scenarios, SMA treats images as a ``contextual signal'' for generating descriptive text. By jointly perturbing both images and text, it achieves cross-modal attribution capabilities. For example, if a word's attribution score remains stable under varying image noise levels, it likely originates from the LLM; if the attribution fluctuates significantly with image perturbations, it is more likely derived from the retrieved visual content.

\begin{figure*}[htbp]
    \centering
    \includegraphics[width=\textwidth]{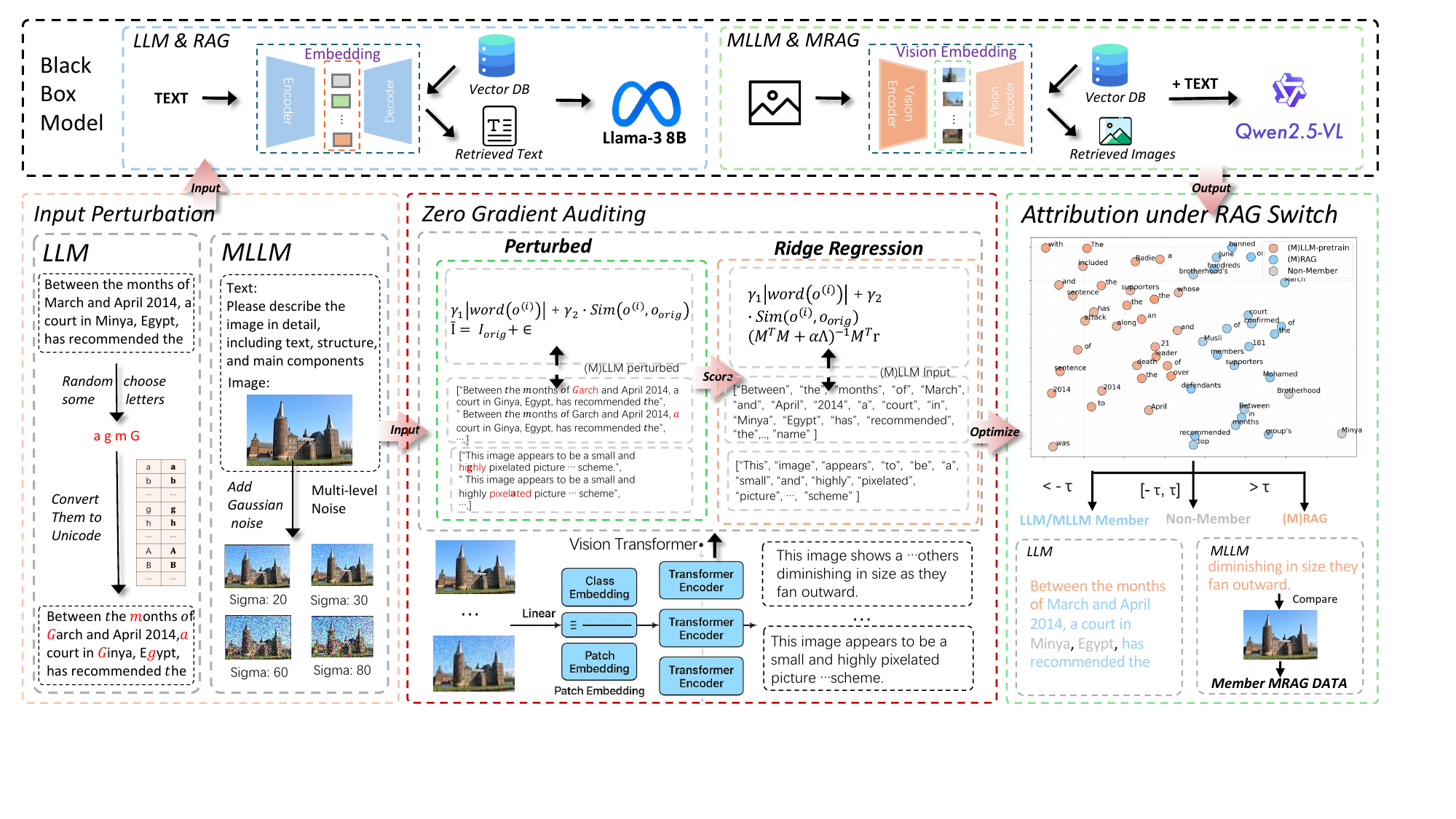}
    \caption{Overview of the SMA framework for LLMs and MLLMs under RAG and MRAG settings, illustrating the token-based attribution pipeline}
    \label{fig:Architec}
\end{figure*}

The overall process of SMA is illustrated in Figure~\ref{fig:Architec}, which includes three stages: structured perturbation of text and image inputs, querying the black-box model and collecting outputs, and performing attribution modeling via ridge regression. The full procedure is detailed in Algorithm~\ref{alg:sma}, which shows how to estimate the contribution of each input token to the final output through perturbation sampling, response scoring, and regression analysis, thereby enabling automated auditing and tracing of knowledge sources.

\begin{algorithm}[H]
\caption{Source Member Attribution (SMA)}
\label{alg:sma}
\begin{algorithmic}[1]
\Require Input $\mathbf{x}$ (text or multimodal), model $f_\theta$, number of perturbations $N$, similarity weights $\gamma_1, \gamma_2$, regularization $\alpha$
\Ensure Token-level attribution score vector $\boldsymbol{\beta} \in \mathbb{R}^L$

\State Extract textual tokens: $\mathbf{x}^{\text{text}} = \{w_1, w_2, \dots, w_L\}$
\State Obtain original model output: $\hat{y} = f_\theta(\mathbf{x})$
\For{$i = 1$ to $N$}
    \State Sample binary mask vector $\mathbf{m}^{(i)} \in \{0,1\}^L$ uniformly
    \If{$\sum_j m_j^{(i)} = 0$}
        \State Set one random $m_j^{(i)} \leftarrow 1$
    \EndIf
    \State Apply character-level or Unicode perturbation to masked tokens to form $\tilde{\mathbf{x}}^{(i)}$
    \State Query the model: $\hat{y}^{(i)} = f_\theta(\tilde{\mathbf{x}}^{(i)})$
    \State Compute response score:
    \[
    \mathbf{r}^{(i)} = \gamma_{1}\,
\frac{\operatorname{Len}\!\bigl(\hat{\mathbf{y}}^{(i)}\bigr)}
     {\operatorname{Len}(\tilde{\mathbf{x}}^{(i)})}
\;+\;
\gamma_{2}\,
\operatorname{Sim}\!\bigl(\hat{\mathbf{y}}^{(i)},\,\tilde{\mathbf{x}}^{(i)}\bigr)
    \]
\EndFor
\State Stack $\{\mathbf{m}^{(i)}\}_{i=1}^N$ into binary mask matrix $M \in \{0,1\}^{N \times L}$

\State Form response vector: $\mathbf{y} = (y_1, y_2, \dots, y_N)^\top$

\State Solve ridge regression:
\[
\boldsymbol{\beta} = (\mathbf{M}^\top \mathbf{M} + \alpha\boldsymbol{\Lambda}\bigr)^{-1} \mathbf{M}^\top \mathbf{r}
\]

\For{each word $j = 1$ to $L$}
    \If{$\boldsymbol{\beta}_j \geq \tau$}
        \State Label word $w_j$ as \emph{Pretrained Member}
    \Else
        \State Temporarily label word $w_j$ as \emph{Retrieved/Non Member}
    \EndIf
\EndFor

\For{each word $j$ labeled as \emph{Retrieved/Non Member}}
    \If{$\tau_1 \leq \text{Diff}(\boldsymbol{\beta}_j) < \tau_2$}
        \State Re-label word $w_j$ as \emph{Non-Member}
    \Else
        \State Re-label word $w_j$ as \emph{Retrieved Member}
    \EndIf
\EndFor

\State \Return Token belonging
\end{algorithmic}
\end{algorithm}

\section{Experiment and Evaluation}
\label{sec: Experiment and Evaluation}

In this section, we present a comprehensive evaluation of our proposed zero-gradient attribution-based attack framework. Our goal is to assess its effectiveness in identifying the provenance of generated content—distinguishing whether specific output tokens originate from retrieved inputs (RAG/MRAG) or from the internal knowledge of the underlying LLM or MLLM.

\subsection{Experiment setup}

\textbf{Datasets.} We use two datasets ragbench and PubMedQA~\cite{jin2019pubmedqadatasetbiomedicalresearch} for rag storage. For MIA comparison, we used WikiMIA~\cite{shi2023detecting} and WikiMIA-24~\cite{fu2024membership} datasets. We also use the two databases VL-MIA-image~\cite{Srinivasan_2021} and the Wikipedia image datasets for MRAG evaluation.

\textbf{Baseline.}  For textual RAG evaluation, we employed LLaMA-2 7B~\cite{touvron2023llama2openfoundation}, LLaMA-3.1 8B~\cite{grattafiori2024llama3herdmodels}, and Qwen2.5 7B~\cite{qwen2.5} language models. Retrieval contexts were generated using three embedding methods: All-MiniLM-L6-v2, bge-large-en-v1.5~\cite{li2024makingtextembeddersfewshot}, and gte-Qwen2-1.5B-instruct~\cite{li2023generaltextembeddingsmultistage}.

For MRAG, we evaluated using the multi-model Qwen2.5-VL-7B model~\cite{Qwen2.5-VL} with the CLIP vit-base-patch32 image embedding model. In addition, we utilized commercial large language models, including ChatGPT-4o mini~\cite{openai2024gpt4technicalreport}, ChatGPT-4.1 mini~\cite{openai2024gpt4technicalreport}, and Gemini-2.5 flash~\cite{comanici2025gemini25pushingfrontier}, as our black-box model baselines.

\textbf{Evaluation metrics} We utilize two primary metrics, accuracy and coverage. \textbf{Accuracy (ACC).} Defined as the cosine similarity between embeddings of attributed RAG-derived outputs and original retrieval results, measuring the fidelity of our attribution:
\begin{equation}
    \text{ACC} = \text{cos}\left(\text{Enc}(o_{\text{rec}}), \text{Enc}(o_{\text{orig}})\right)
\end{equation}

\textbf{Coverage.} Calculated as the ratio of identified RAG-derived tokens to total tokens retrieved by RAG systems, indicating how comprehensively our attribution captures retrieval content. \textbf{FPR} uses to quantify the proportion of data that are incorrectly identified as members.  \textbf{Member Data Accuracy (MDA).} Defined $N_{\mathrm{correct}}^{\mathrm{member}}$ is the number of member data correctly identified as members, and $N_{\mathrm{total}}^{\mathrm{member}}$ is the total number of member data samples.

\begin{equation}
\mathrm{MDA} = \frac{N_{\mathrm{correct}}^{\mathrm{member}}}{N_{\mathrm{total}}^{\mathrm{member}}}
\end{equation}

\textbf{Non-member Data Accuracy (NMDA).} Defined $N_{\mathrm{correct}}^{\mathrm{non\text{-}member}}$ is the number of non-member data correctly identified as non-members, and $N_{\mathrm{total}}^{\mathrm{non\text{-}member}}$ is the total number of non-member data samples.

\begin{equation}
\mathrm{NMDA} = \frac{N_{\mathrm{correct}}^{\mathrm{non\text{-}member}}}{N_{\mathrm{total}}^{\mathrm{non\text{-}member}}}
\end{equation}

\subsection{Experimental Results}

\subsubsection{Evaluation of RAG\&LLM System Performance Across Models}
To evaluate the generalizability of SMA accross different RAG embedding models and various LLMs, we combine three mainstream foundation LLMs including LLaMA-3.1 8B, Qwen2.5 7B, and LLaMA-2 7B, with three different RAG embedding models, such as all-MiniLM-L6-v2, bge-large-en-v1.5 and te-Qwen2-1.5B-instruct. Table~\ref{tab:rag-llm-comparison}  reports two key metrics, ACC and coverage, for each combination across the RB and PMQA datasets, under the retrieval $\text{top}_{\text{k}}$ setting of 3. We can draw three main conclusions: \ding{182} The accuracy of SMA can be influenced by the choice of embedding methods. For example, the bge-large-en-v1.5 embedding boosts LLaMA-3.1 8B's ACC on PMQA to 0.9172, while the same model with all-MiniLM-L6-v2 on RB achieves only 0.7230. \ding{183} SMA The type of LLMs can also affect the performance of SMA. \ding{184}There exists a trade-off between accuracy and coverage across different models and embeddings. For instance, although Qwen2.5 7B attains a high accuracy of 0.8288 using all-MiniLM-L6-v2 on RB, its coverage drops to 0.1724, compared to LLaMA-3.1 8B, which maintains a more balanced profile with ACC 0.7230 and coverage 0.8095 under the same setting.

In summary, these results highlight the importance of careful selection and tuning of both embedding models and LLMs to optimize the overall performance of RAG-based systems, as both accuracy and coverage are sensitive to these choices.

\begin{table*}[htbp]
  \centering
  \caption{Performance Comparison of RAG Embedding Models Combined with LLMs  with ${top_k}$=3}
  \label{tab:rag-llm-comparison}
  \setlength{\tabcolsep}{5pt}
  \small
  \begin{tabular}{ccccccc}
    \hline
    \toprule
    \multirow{3 }{*}{RAG\&LLM Models}
      & \multicolumn{2}{c}{LLaMA-3.1 8B}
      & \multicolumn{2}{c}{Qwen2.5 7B}
      & \multicolumn{2}{c}{LLaMA-2 7B} \\
    \cmidrule(lr){2-3} \cmidrule(lr){4-5} \cmidrule(lr){6-7}
      & {ACC} & {Coverage}
      & {ACC} & {Coverage}
      & {ACC} & {Coverage} \\
    \midrule
    \rowcolor{gray!10} RAG(all-MiniLM-L6-v2)+RB
      & 0.7230 & 0.8095 & 0.8288 & 0.1724 & 0.8030 & 0.6667 \\
    RAG(all-MiniLM-L6-v2)+PMQA
      & 0.7366 & 0.8000 & 0.6700 & 0.5714 & 0.7902 & 0.3500 \\
    \rowcolor{gray!10} RAG(bge-large-en-v1.5)+RB
      & 0.9036 & 0.4500 & 0.6026 & 0.5714 & 0.7483 & 0.5000 \\
    RAG(bge-large-en-v1.5)+PMQA
      & 0.9172 & 0.2619 & 0.6691 & 0.5700 & 0.8445 & 0.5714 \\
    \rowcolor{gray!10} RAG(gte-Qwen2-1.5B-instruct)+RB
      & 0.6083 & 0.3750 & 0.6026 & 0.8000 & 0.8464 & 0.4219 \\
    RAG(gte-Qwen2-1.5B-instruct)+PMQA
      & 0.8369 & 0.6842 & 0.5311 & 0.4000 & 0.9137 & 0.3409 \\
    \bottomrule
  \end{tabular}

  \vspace{0.2cm}
  \footnotesize
  \textbf{Note}: ACC (Accuracy) indicates the similarity between the outputs from membership inference attack (MIA) and the original RAG retrieval results. Coverage represents the proportion of matched RAG data relative to all data retrieved by the RAG system.
\end{table*}

\begin{figure}[htbp]
  \centering
  \begin{subfigure}{0.9\linewidth}
    \centering
    \includegraphics[width=\linewidth]{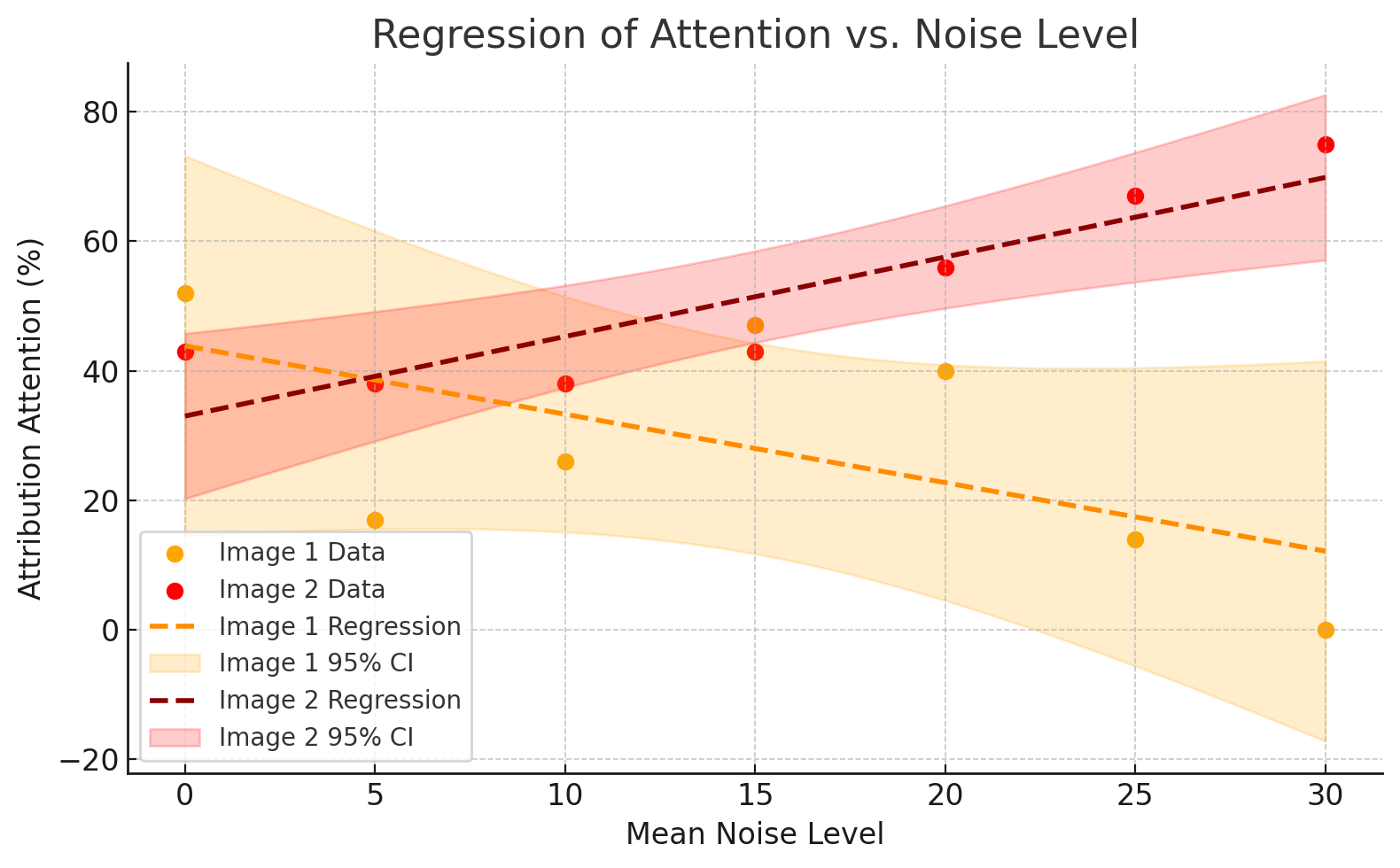}
    \caption{White Box Noise Comparison on attention and Gaussian noise}
    \label{fig:white_box_noise}
  \end{subfigure}
  \hfill
  \begin{subfigure}{0.9\linewidth}
    \centering
    \includegraphics[width=\linewidth]{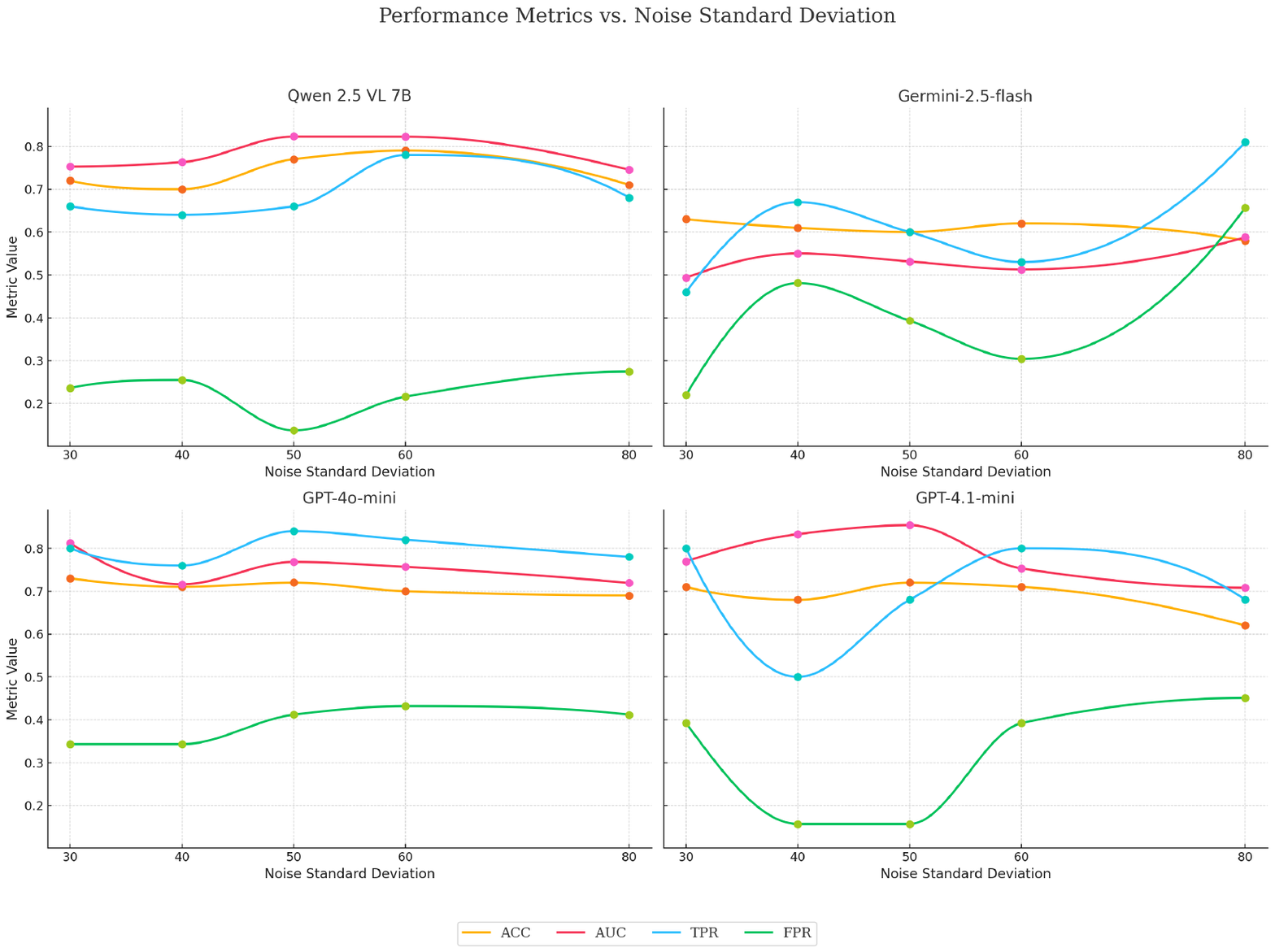}
    \caption{Effect of Gaussian noise on MRAG\&MLLM system performance metrics}
    \label{fig:noise_mllm}
  \end{subfigure}
  \caption{Attribution attention comparison under different noise levels}
  \label{fig:attention_compare}
\end{figure}

In \textbf{MRAG\&MLLM System}. When adding noise, after white-box attribution experiments with MLLMs, Fig.~\ref{fig:white_box_noise} shows that under same prompts and model hyperparameter, introducing noise to an image leads the MLLM to focus more attention on the perturbed image. Meanwhile, results from MRAG experiments demonstrate that adding noise to the image does not affect the MRAG retrieval results. Based on these findings, we first obtain the MLLM inference outputs using the original image paired with the image retrieved by MRAG, and then repeat the process using the noised image. By constructing prompts with the noisy image, we guide the MLLM’s attention toward the targeted image, thereby extracting descriptive content about the other image. We then apply attribution methods to identify which parts of the LLM output originate from the image, and compare these outputs with the original image descriptions to score and evaluate the attribution results. Fig.~\ref{fig:noise_mllm} illustrates the effects of varying levels of Gaussian noise in four different models which represented by different std values introduced to images in MRAG\& MLLM System. We observe how noise influences performance metrics.

As shown in the figure, moderate Gaussian noise which std values around 50 to 60, enhancing the performance of the MLLM, particularly reflected in increased ACC and AUC scores. This improvement indicates that moderate noise effectively directs the model’s attention towards the perturbed images, thus enhancing its capability for distinguishing sensitive visual information, and subsequently improving True Positive Rate. However, excessively high noise at the std value of 80 negatively impacts model performance across all metrics, likely due to significant image distortion that hinders the model’s ability to accurately interpret visual content.

These findings suggest that carefully calibrated noise perturbations can substantially improve the attribution accuracy and robustness of MRAG systems in membership inference attacks, highlighting the importance of optimal noise management for practical deployments.

\subsubsection{Comparison with SoTA method}
We conducted extensive experiments to evaluate the effectiveness of our proposed MIA method in scenarios involving RAG. Specifically, we compare our method, SMA, against five baseline methods, PETEL~\cite{heLabelOnlyMembershipInference2025}, Mink++~\cite{zhang2025mink}, Min-K\% PROB ${(K=10,20,30)}$~\cite{shi2024detecting}, under the six LLM models on two distinct datasets.

\begin{table*}[htbp]
    \centering
    \caption{Performance comparison of MIA methods under RAG settings across six LLM models}
    \label{tab:mia_comparison}
    \begin{tabular}{cccccccc}
        \toprule
        \multirow{2}{*}{Model} & \multirow{2}{*}{Method} & \multicolumn{2}{c}{WikiMIA~\cite{shi2023detecting}} & \multicolumn{2}{c}{WikiMIA-24~\cite{fu2024membership}}  \\
        \cmidrule(r){3-4} \cmidrule(r){5-6}
        & & ACC & Coverage & ACC & Coverage  \\
        \midrule
        \multirow{6}{*}{LLaMA-2 7B}
        & PETEL~\cite{heLabelOnlyMembershipInference2025}   & 0.5230  & 0.5230  & 0.5741 & 0.5741 \\
        & Mink++~\cite{zhang2025mink}  & 0.5190  & 0.5190  & 0.5010   & 0.5010  \\
        & MIN-K\%PROB (K=10\%)~\cite{shi2024detecting} & 0.5032 & 0.5032 & 0.5370 & 0.5370  \\
        & MIN-K\%PROB (K=20\%)~\cite{shi2024detecting} & 0.5015 & 0.5015 & 0.5322& 0.5322\\
        & MIN-K\%PROB (K=30\%)~\cite{shi2024detecting} & 0.5029 & 0.5029 & 0.5314 &  0.5314  \\
        & SMA     & \textbf{0.8624} & \textbf{0.5882} & \textbf{0.6730} & \textbf{0.6000}  \\
        \midrule
        \multirow{6}{*}{LLaMA-3.1 8B}
        & PETEL~\cite{heLabelOnlyMembershipInference2025}   & 0.5332 & 0.5332 & 0.6280 & 0.6280  \\
        & Mink++~\cite{zhang2025mink}  & 0.5200  & 0.5200  & 0.5960  & 0.5960 \\
        & MIN-K\%PROB (K=10\%)~\cite{shi2024detecting} & 0.5126 & 0.5126 &  0.5556 &  0.5556 \\
        & MIN-K\%PROB (K=20\%)~\cite{shi2024detecting} & 0.5074 & 0.5074 &  0.5378 &  0.5378  \\
        & MIN-K\%PROB (K=30\%)~\cite{shi2024detecting} & 0.5045 & 0.5045 &  0.5531 &  0.5531 \\
        & SMA     & \textbf{0.6351} & \textbf{0.7895}  & \textbf{0.7233} & 
        \textbf{0.6667} \\
        \midrule
        \multirow{6}{*}{Qwen2.5 7B}
        & PETEL~\cite{heLabelOnlyMembershipInference2025}   & 0.5088 & 0.5088 & 0.4968 & 0.4968\\
        & Mink++~\cite{zhang2025mink}  & 0.5890  & 0.5890   & 0.5540  & 0.5540  \\
        & MIN-K\%PROB (K=10\%)~\cite{shi2024detecting} & 0.5000 & 0.5000 &  0.5918 &  0.5918 \\
        & MIN-K\%PROB (K=20\%)~\cite{shi2024detecting} & 0.5072 & 0.5072 &  0.5676 &  0.5676 \\
        & MIN-K\%PROB (K=30\%)~\cite{shi2024detecting} & 0.5035 & 0.5035 &  0.5523 &  0.5523 \\
        & SMA     & \textbf{0.6484} & \textbf{0.6700}  & \textbf{0.5705} & 
        \textbf{0.7500} \\
        \midrule
        \multirow{6}{*}{ChatGPT 4.1-mini}
        & PETEL~\cite{heLabelOnlyMembershipInference2025}   & 0.5370 & 0.5370 & 0.5384  & 0.5384  \\
        & Mink++~\cite{zhang2025mink}  & 0.3621  & 0.3621  & 0.4909  & 0.4909 \\
        & MIN-K\%PROB (K=10\%)~\cite{shi2024detecting} & 0.5025  & 0.5025  & 0.5306  & 0.5306 \\
        & MIN-K\%PROB (K=20\%)~\cite{shi2024detecting} & 0.5023  & 0.5023  & 0.5209  & 0.5209 \\
        & MIN-K\%PROB (K=30\%)~\cite{shi2024detecting} & 0.5024  & 0.5024  & 0.5185  & 0.5185 \\
        & SMA     & \textbf{0.7489} & \textbf{0.5455}  & \textbf{0.7394} & 
        \textbf{0.5400} \\
        \midrule
        \multirow{6}{*}{ChatGPT 4o-mini}
        & PETEL~\cite{heLabelOnlyMembershipInference2025}   & 0.5625 & 0.5625 & 0.5537  & 0.5537  \\
        & Mink++~\cite{zhang2025mink}  & 0.4290  & 0.4290  & 0.4733  & 0.4733 \\
        & MIN-K\%PROB (K=10\%)~\cite{shi2024detecting} & 0.5700  & 0.5700  & 0.5411  & 0.5411 \\
        & MIN-K\%PROB (K=20\%)~\cite{shi2024detecting} & 0.5789  & 0.5789  & 0.5580  & 0.5580 \\
        & MIN-K\%PROB (K=30\%)~\cite{shi2024detecting} & 0.5749  & 0.5749  & 0.5797  & 0.5797 \\
        & SMA     & \textbf{0.6276} & \textbf{0.6273}  & \textbf{0.6839} & 
        \textbf{0.6012}\\
        \midrule
        \multirow{6}{*}{Gemini 2.5-flash}
        & PETEL~\cite{heLabelOnlyMembershipInference2025}   & 0.5614 & 0.5614 & 0.5583  & 0.5583 \\
        & Mink++~\cite{zhang2025mink}  & NaN  & NaN  & NaN  & NaN \\
        & MIN-K\%PROB (K=10\%)~\cite{shi2024detecting} & NaN  & NaN  & NaN  & NaN \\
        & MIN-K\%PROB (K=20\%)~\cite{shi2024detecting} & NaN  & NaN  & NaN  & NaN \\
        & MIN-K\%PROB (K=30\%)~\cite{shi2024detecting} & NaN  & NaN  & NaN  & NaN \\
        & SMA     & \textbf{0.5828} & \textbf{0.6000}  & \textbf{0.6137} & 
        \textbf{0.5669} \\
        \bottomrule
    \end{tabular}
\end{table*}

Table~\ref{tab:mia_comparison} presents a comprehensive evaluation of different MIA methods—PETEL, Mink++, Min-K\%PROB ${(K=10,20,30)}$, and our proposed SMA framework—across six LLM models and two benchmark datasets under RAG settings. The table reports ACC and coverage for each method and model combination. It is evident that the SMA framework consistently outperforms the baselines by a substantial margin. For instance, on the WikiMIA dataset with LLaMA-2 7B, SMA achieves an accuracy of 0.8624 and a coverage of 0.5882, while others remain below 0.53 on both metrics. Similarly, with LLaMA-3.1 8B on WikiMIA-24, SMA attains 0.7233 accuracy, significantly surpassing the highest baseline of 0.6280. Notably, SMA also demonstrates strong coverage on Qwen2.5 7B, achieving up to 0.7500. These results highlight the robustness and effectiveness of the SMA framework, establishing a new state-of-the-art in membership auditing for retrieval-augmented LLMs, especially under black-box constraints. It is worth noting that existing methods such as PETEL and Mink++ exhibit inherent incompatibility with commercial API-based language models like ChatGPT-4o mini, ChatGPT-4.1 mini, and Gemini-2.5 flash. Specifically, these baseline approaches require direct access to internal parameters, such as gradient or singal token, to effectively perform membership inference. However, prevalent commercial APIs typically restrict output to plain textual responses, withholding granular internal details. Consequently, PETEL, Mink++, and similar methods relying on such internal model specifics are substantially constrained and fail to adapt efficiently to these black-box commercial environments. For PETEL and Min-K\%PROB, we implemented minimal modifications to accommodate commercial APIs. Additionally, the Gemini API does not return the parameters required by Min-K\%PROB and Mink++, further limiting its applicability. In contrast, our proposed SMA framework, operating solely on externally observable input-output perturbations, circumvents these limitations, making it uniquely suited and adaptable for auditing membership inference in contemporary commercial API models. By attributing input data and extracting relevant information from the RAG database when identified, our method notably outperforms the baseline methods.

We also conducted comprehensive evaluations to compare our SMA framework against existing MRAG based MIA methods, specifically VLMA~\cite{liMembershipInferenceAttacks2024} and vlm\_mia~\cite{hu2025membershipinferenceattacksvisionlanguage}, using the VL-MIA-images dataset with the base model Qwen2.5-VL 7B, ChatGPT-4o mini, ChatGPT-4.1 mini, and Gemini-2.5 flash. Table~\ref{tab:mia_comparison} shows the performance of different MIA methods—VLMA, VLM\_MIA, and our SMA framework—under the MRAG scenario with ${top_k}$, evaluated using multiple metrics including ACC, AUC, TPR and FPR. The results clearly demonstrate that the SMA framework achieves superior overall performance, setting new benchmarks for both accuracy and reliability. Specifically, SMA attains the highest accuracy (0.7900) and AUC (0.8227) in Qwen2.5 VL 7B, while also achieving the lowest false positive rate (FPR) of 0.0785 among all methods. In comparison, VLMA and VLM\_MIA not only fall behind in accuracy and AUC, but also suffer from substantially higher FPR values, with VLM\_MIA's FPR reaching 0.9968. These results highlight the robust capability of the SMA framework to effectively audit membership information in multimodal RAG systems, outperforming existing methods across nearly all evaluation dimensions.

Our experiments revealed that VLMA experiences a reduction in attack accuracy when MRAG is introduced, compared to its performance reported originally. Additionally, our observations indicate that as the \texttt{${top_k}$} parameter in the MRAG system increases, the ACC of the membership inference attack correspondingly decreases.

The attack principle employed in our MRAG-based MIA method closely follows the approach used in the RAG\&LLM system scenario. Specifically, we separate images retrieved by MRAG from those provided by the user and apply attribution techniques to isolate content derived from the pre-trained MLLM. By assessing whether the user-provided images align with the target images, we effectively conduct the MIA. The results clearly demonstrate the superiority of our SMA method in the presence of MRAG. Specifically, our method achieved an ACC of 0.79, higher than VLMA's 0.5983 and vlm\_mia's 0.4985. 

\begin{table*}[htbp]
  \centering
  \small
  \caption{Performance comparison of MIA methods under RAG settings across LLMs}
  \label{tab:mia_comparison}
  \begin{tabular}{cccccc}
    \toprule
    Model& Method & ACC & AUC & TPR &  FPR \\
    \midrule
    \multirow{3}{*}{Qwen2.5 VL 7B} 
    & VLM\_MIA~\cite{hu2025membershipinferenceattacksvisionlanguage} & 0.4311 & 0.5208 & 0.6100 & 0.5966 \\
    & VLMA~\cite{liMembershipInferenceAttacks2024} & 0.4985 & 0.4303 & 0.9968 & 0.9998 \\
    & SMA & \textbf{0.7900} & \textbf{0.8227} & 0.6400 & \textbf{0.0785} \\
    \midrule
    \multirow{3}{*}{ChatGPT 4o-mini} 
    & VLM\_MIA~\cite{hu2025membershipinferenceattacksvisionlanguage} & 0.5000 & 0.2476 & 0.2480 & 1.0000 \\
    & VLMA~\cite{liMembershipInferenceAttacks2024} & 0.5865 & 0.5160 & 0.5900 & 1.0000 \\
    & SMA & \textbf{0.7700} & \textbf{0.8541} & \textbf{0.6800} & \textbf{0.1569} \\
    \midrule
    \multirow{3}{*}{ChatGPT 4.1-mini} 
    & VLM\_MIA~\cite{hu2025membershipinferenceattacksvisionlanguage} & 0.5000 & 0.0237 & 0.0000 & 1.0000 \\
    & VLMA~\cite{liMembershipInferenceAttacks2024} & 0.5176 & 0.2917 & 0.1000 & 1.0000  \\
    & SMA & \textbf{0.7300} & \textbf{0.8123} & \textbf{0.8000}  & \textbf{0.3529} \\
    \midrule
    \multirow{3}{*}{Gemini 2.5 flash} 
    & VLM\_MIA~\cite{hu2025membershipinferenceattacksvisionlanguage} & 0.6193 & 0.8945 &  0.2628 & 0.9948 \\
    & VLMA~\cite{liMembershipInferenceAttacks2024} & 0.5449 & 0.4503 & 0.4000 & 0.8750 \\
    & SMA & \textbf{0.6300} & \textbf{0.4937} & \textbf{0.6000} & \textbf{0.2157} \\
    \bottomrule
  \end{tabular}
\end{table*}

In both retrieval‐augmented settings, whether text‐only RAG or MRAG. Our SMA consistently outperforms prior membership inference methods. By distinguishing content sourced from the external database versus the model’s own pre-training, SMA maintains high accuracy and coverage even when RAG or MRAG is enabled, whereas existing techniques like PETEL, Mink++, and VLMA suffer substantial drops in performance once the retrieval component is active.

\subsection{Ablation Studies}

To demonstrate the contribution of each component in our attack framework, we conducted a comprehensive ablation study across both the RAG\&LLM and MRAG\&MLLM systems. We have decomposed our approach into (i) a basic zero-gradient attribution method, and (ii) the combination of noise injection with zero-gradient attribution. This experimental setup allows us to systematically assess the necessity and effect of noise perturbation for optimizing our method. With the bge-large-en-v1.5 embedding model, ${top_k}$ set to 8 and the ragbench dataset.

\begin{table}[htbp]
    \centering
    \caption{Ablation Study for LLMs and MLLM for each singal method}
    \label{tab:ablation_compact}
    \renewcommand{\arraystretch}{1.05}
    \begin{tabular}{cccc}
        \midrule
        Type & Model & Attr+ZeroGrad & +Noise \\
        \midrule
         \multirow{3}{*}{LLM}
        & LLaMA-3.1 8B & 0.7666 & \textbf{0.7895} \\
        & Qwen2.5 7B & 0.6033 & \textbf{0.6484} \\
        & ChatGPT-4o mini & 0.5733 & \textbf{0.6276} \\
        \midrule
        \multirow{3}{*}{MLLM}
        & Qwen2.5-VL 7B & 0.6500 & \textbf{0.7900} \\
        & ChatGPT-4o mini & 0.7100 & \textbf{0.7700} \\ 
        & Gemini-2.5 flash & 0.5100  & \textbf{0.6300} \\
        \bottomrule
    \end{tabular}
\end{table}

As shown in Table~\ref{tab:ablation_compact}, it presents the results of ablation experiments, evaluating the accuracy impact of zero-gradient attribution alone (Attr+ZeroGrad) and the addition of noise (+Noise) across both LLM and MLLM models. The table is composed of results for LLaMA-3.1 8B, Qwen2.5 7B and ChatGPT-4o mini(LLMs) as well as Qwen2.5-VL 7B (MLLM). From table, two conclusions can be drawn: First, adding noise consistently improves the accuracy for all models. For example, the accuracy of LLaMA-3 8B increases from 0.7666 to 0.7895, while Qwen2.5 7B rises from 0.6033 to 0.6484, and ChatGPT-4o mini from 0.5733 up to 0.6276. In visual Model, Qwen2.5-VL 7B goes from 0.65 to 0.7900, ChatGPT-4o mini goes from 0.7100 to 0.7700, Gemini-2.5 flash goes from 0.5100 to 0.6300 after noise is added. Second, although the absolute increase varies, the largest gain is observed for Gemini-2.5 flash, where accuracy improves by 0.1200.

Overall, these results demonstrate that combining zero-gradient attribution with noise is highly effective for both language and multimodal models, as evidenced by the substantial improvement in accuracy—for instance, an increase of 0.1200 for Gemini-2.5 flash.

\begin{table}[htbp]
    \centering
    \caption{Comparison with White-box Attribution in the ragbench dataset and ${\text{top}_k}$ set to 8, all-MiniLM-L6-v2 embedding model}
    \label{tab:llm_mllm_ablation}
    \renewcommand{\arraystretch}{1.05}
    \begin{tabular}{ccc}
        \toprule
        Method & LLaMA-3.1 8B & Qwen2.5 7B \\
        \midrule
        \rowcolor{gray!10} Attr+Noise+White-box & 0.8612 & 0.8073 \\
        Attr+Noise+Alt. Model  & 0.7058 & 0.6464 \\
        \rowcolor{gray!10} Attr+Noise+Zero-Grad  & 0.7008 & 0.6657  \\
        \bottomrule
    \end{tabular}
\end{table}

Ablation result across White-Box attribution: Table~\ref{tab:llm_mllm_ablation} shows the comparison result in methods of White-Box Attribution with Noise, White-Box Attrition with Alternative Model and our current method Black-Box zero-gradient Attribution with Noise under two LLM categories. We can know that: First, The White-Box Attribution with Noise is under the desired condition, thus the average Accuracy Score of two LLMs get the 0.8612 and 0.8073. Second, the the alternative models (the two models are exchanged for attribution) reach the 0.7058 and 0.6464. Finally, our current method Black-Box Attribution with Noise get the Accuracy of 0.7008 and 0.6657 by two LLMs.

\begin{figure}
    \centering
    \includegraphics[width=1\linewidth]{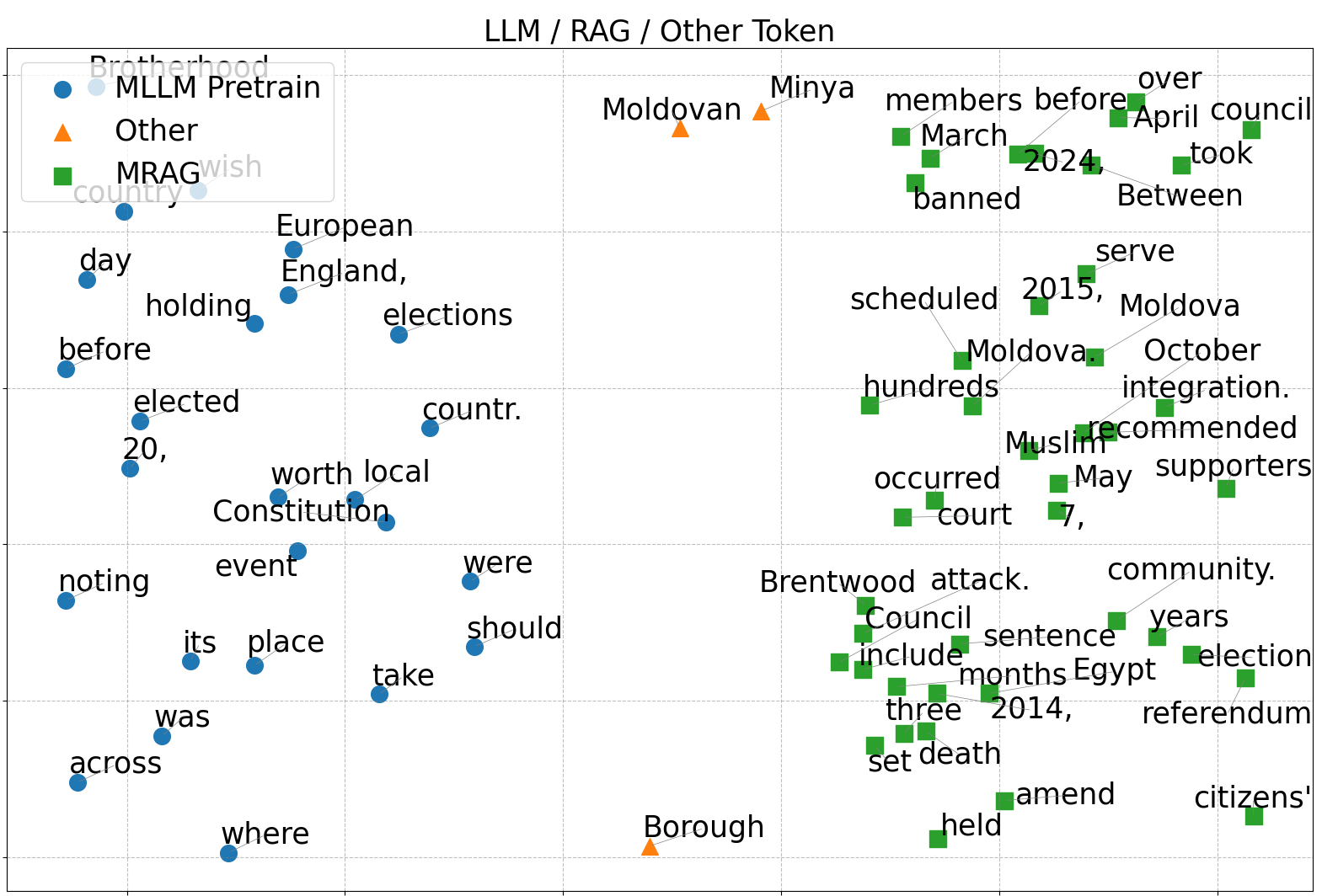}
    \caption{Separation of token between RAG and LLM pre-train based on ${top_k}$=3}
    \label{fig:attr-token}
\end{figure}

In summary, our experiments conclusively demonstrate that noise perturbation significantly enhances MIA accuracy in both text-only and multi-model systems. Moreover, the gap between our semi-black-box method and the white-box upper bound is relatively narrow, highlighting the effectiveness and practicality of our approach.

\begin{figure}
    \centering
    \includegraphics[width=1\linewidth]{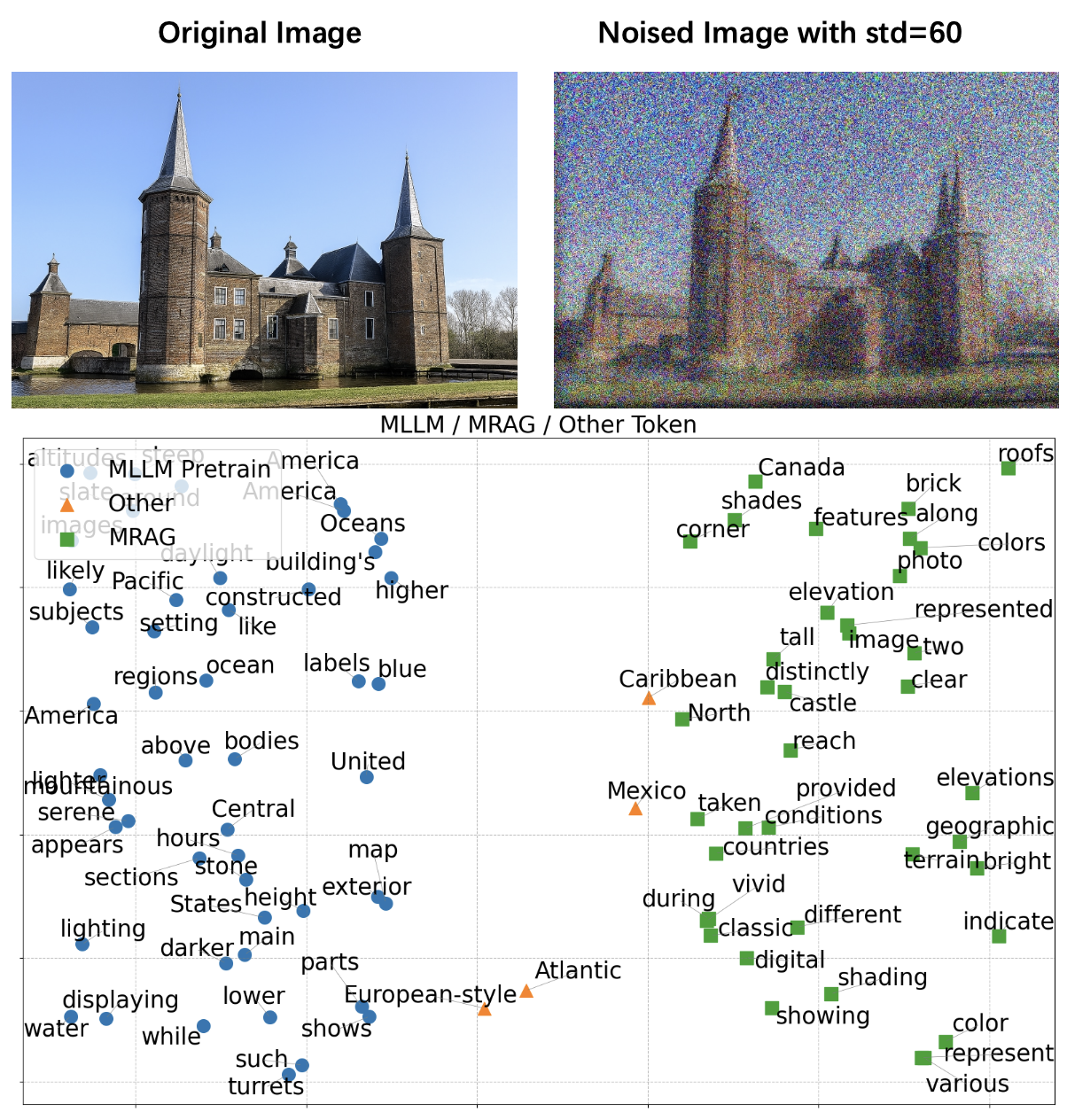}
    \caption{Separation of token between MRAG and MLLM pre-train based on ${top_k}$=1}
    \label{fig:mllm-attr-token}
\end{figure}
\section{Discussion} \label{sec: Discussion}

\textbf{Costing.} One practical consideration of our SMA framework is the inference cost associated with black-box access. Since SMA operates in a RAG black-box setting and relies on perturbation-based zero-gradient attribution, each query potentially requires multiple forward passes to complete attribution across tokens. When accessed via APIs, this leads to increased token consumption and corresponding monetary cost. The total token usage for each inference in SMA can be approximated as $(\text{Token}_{\text{SMA}} = \text{Token}_\text{Output From Taret} + 60)$. The actual cost depends on the maximum token limits and pricing model of the deployed server. Fortunately, with the emergence of low-cost model providers such as DeepSeek, the economic overhead is increasingly manageable—for example, 1M tokens currently cost as little as \$0.07. To further mitigate operational costs and avoid triggering abnormal API usage alerts, a promising future direction is to integrate shadow model inference~\cite{salem2018mlleaksmodeldataindependent}~\cite{carlini2022membershipinferenceattacksprinciples}. This approach would replicate the queried model locally, allowing efficient large-scale offline testing of the SMA framework without incurring external token charges or raising API rate-limiting alarms. Such optimizations could substantially improve the deployment practicality of SMA in high-volume audit scenarios.

\begin{figure}[htbp]
  \centering
  \begin{subfigure}{0.48\textwidth}
    \centering
    \includegraphics[width=\linewidth]{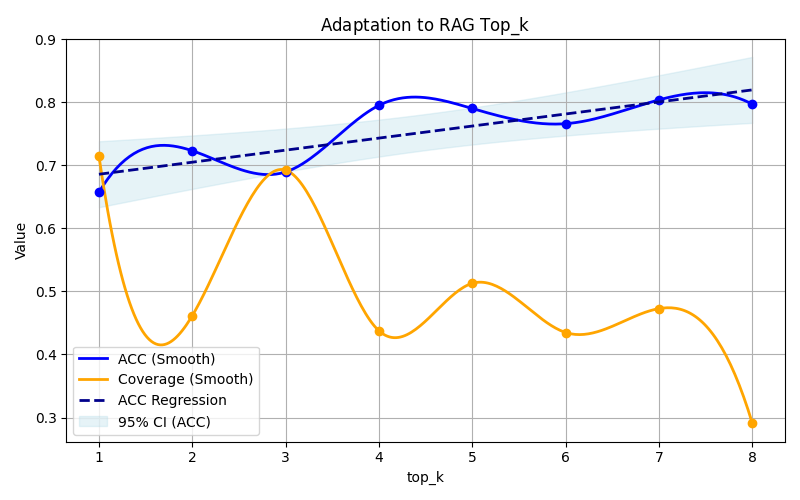}
    \caption{Adaptation to RAG ${top_k}$}
    \label{fig:adapt_topk}
  \end{subfigure}
  \hfill
  \begin{subfigure}{0.48\textwidth}
    \centering
    \includegraphics[width=\linewidth]{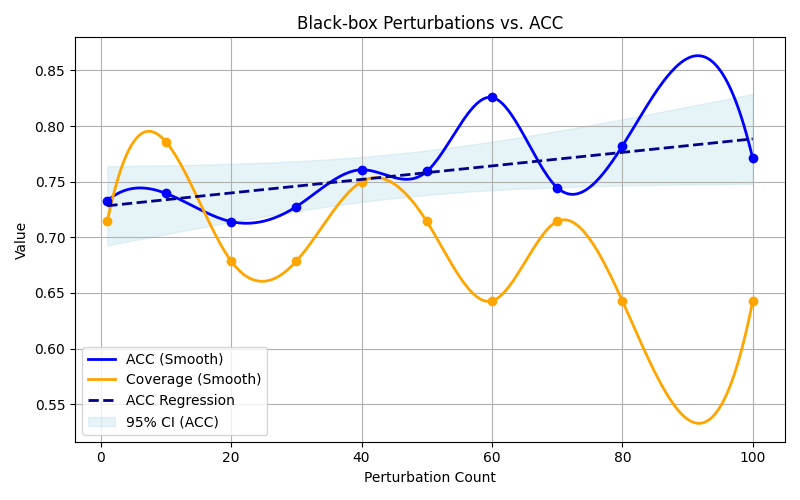}
    \caption{Black‐box Perturbations vs.\ ACC}
    \label{fig:perturb_acc}
  \end{subfigure}
  \caption{(a) Varying RAG retrieval depth shows that SMA sustains high ACC and graceful coverage degradation even as ${top_k}$ grows. (b) Varying the number of zero‐gradient perturbations reveals an optimal budget that maximizes ACC in black‐box MIA.}
  \label{fig:param_sensitivity}
\end{figure}

\textbf{Parameter Sensitivity.} Another aspect of our analysis involves understanding how the SMA attack behaves under different parameter settings in both RAG\&LLM and MRAG\&MLLM System scenarios. In Figure~\ref{fig:param_sensitivity}, we evaluate two key hyperparameters, the RAG number of results ${top_k}$, and the number of perturbation queries used in zero-gradient black-box attribution. In Fig.\ref{fig:adapt_topk} , we vary ${top_k}$ from 1 to 8, which controls how many documents are retrieved from the RAG system and appended to the prompt. Despite the increasing uncertainty introduced by adding more retrieved passages, our SMA method maintains a consistently high ACC across different ${top_k}$ values. This suggests that SMA is robust to varying retrieval depths and can effectively filter out non-member content from diverse inputs. Fig.~\ref{fig:perturb_acc} explores the effect of different perturbation counts, ranging from 1 to 100. Perturbations influence the precision of attribution in zero-gradient settings: too few queries may yield noisy gradients, while excessive queries increase computational load. Our results show that SMA achieves stable and high accuracy around 60--80 perturbations, with diminishing returns beyond that point. The number of perturbations can thus be flexibly tuned based on the attacker’s computational budget and accuracy requirement.

\textbf{Limitations. } While SMA demonstrates strong performance under RAG scenarios, it has several practical limitations. First, as a strictly semi-black-box technique, SMA is constrained by the language model’s maximum token limit and sampling temperature. Specifically, if the temperature is set above typical operational ranges such as temperature bigger than 5.0 generated outputs can become overly stochastic, degrading attribution consistency and attack accuracy. Similarly, overly restrictive max\_tokens settings truncate the context window, preventing sufficient perturbation analysis and impairing the attack’s success rate. Second, SMA’s reliance on repeated perturbation-based queries leads to high CPU and API usage. Conducting dozens to hundreds of semi-black-box attribution passes incurs significant computational overhead and latency, resulting in longer overall execution times compared to white-box or single-shot methods. Future work may explore optimizations such as batched perturbations or early stopping criteria to alleviate these resource constraints.

\section{Conclusion} \label{sec: Conclusion}

In this paper, we propose SMA (Source-aware Membership Auditing), the first membership auditing framework with source attribution capability, to determine whether the leaked content originates from the model's pre-training corpus or an external retrieved data source in a black-box setting. The core of SMA consists of two key designs: (1) response sensitivity analysis based on lightweight input perturbations for eliciting differential variations across sources; and (2) a zero-gradient scoring mechanism that does not require gradient information and is suitable for black-box environments in deployed systems. To support multi-model inputs, SMA also introduces a unified cross-modal attribution mechanism, which for the first time enables cross-modal auditing of member leakage in MRAG systems. The experimental results validate its significant advantages in textual and multi-model tasks, providing a practical tool for data compliance and privacy auditing in generative systems.

\bibliographystyle{unsrt}
\bibliography{ref}

\begin{figure*}
    \centering
    \includegraphics[width=1\linewidth]{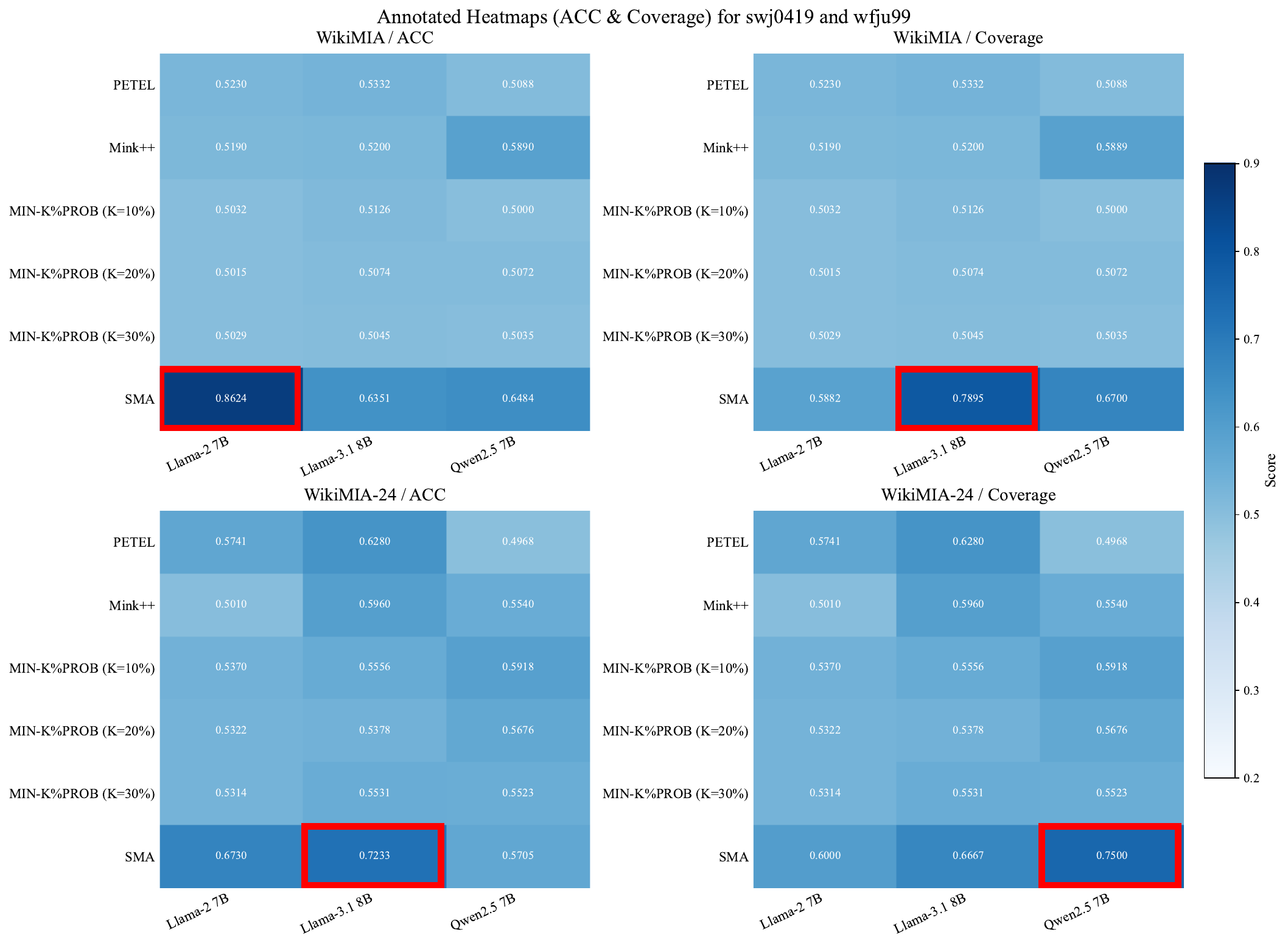}
    \caption{Heat map of MIA methods under RAG settings across six LLM models}
    \label{fig:heap_map_mia}
\end{figure*}

\appendix
Fig.~\ref{fig:heap_map_mia} provides supplementary heatmaps for the LLM experiments described in the main text, visually comparing the performance of different MIA methods across multiple models and evaluation metrics.
\section{Show}

\end{document}